\documentclass[final, 12pt]{elsarticle}
\usepackage{physics}
\usepackage[shortlabels]{enumitem}
\usepackage{mathtools}
\usepackage{amsthm}
\usepackage[utf8]{inputenc}
\usepackage{graphicx}
\graphicspath{ {images/} }
\usepackage[english]{babel}
\usepackage[margin=1in]{geometry}
\usepackage{url}
\usepackage{amsmath, nccmath}
\usepackage{float}
\usepackage{caption}
\usepackage{comment}
\usepackage{lscape}
\usepackage{afterpage}
\usepackage{bookmark}
\newcommand{\diff}{\,\text{d}}

\newcommand\restr[2]{{
		\left.\kern-\nulldelimiterspace 
		#1 
		\vphantom{\big|} 
		\right|_{#2} 
}}

\newcommand{\tnorm}{\mathcal{T}}
\newcommand{\imp}{\mathcal{I}}
\newcommand{\pwset}{\mathcal{P}}
\newcommand{\fpwset}{\widetilde{\mathcal{P}}}

\usepackage[utf8]{inputenc}
\usepackage{lmodern}
\usepackage{mathtools}
\newtheorem{thm}{Theorem}[section] 
\newtheorem{prop}[thm]{Proposition}
\newtheorem{lem}[thm]{Lemma}
\newtheorem{cor}[thm]{Corollary}

\newtheorem{defn}[thm]{Definition} 
\newtheorem{example}[thm]{Example}
\usepackage{amssymb}
\usepackage{esint}
\usepackage{hyperref}
\begin{document}
\begin{frontmatter}

    \title{On the Granular Representation of Fuzzy Quantifier-Based Fuzzy Rough Sets}
    \author[mymainaddress]{Adnan Theerens\corref{mycorrespondingauthor}}
    \cortext[mycorrespondingauthor]{Corresponding author}
    \ead{adnan.theerens@ugent.be}
    \author[mymainaddress]{Chris Cornelis}
    \ead{chris.cornelis@ugent.be}
    \address[mymainaddress]{Computational Web Intelligence, Department of Applied Mathematics, Computer Science and Statistics, Ghent University, Ghent, Belgium}

    \begin{abstract}
        Rough set theory is a well-known mathematical framework that can deal with inconsistent data by providing lower and upper approximations of concepts. A prominent property of these approximations is their granular representation: that is, they can be written as unions of simple sets, called granules. The latter can be identified with ``if…, then… '' rules, which form the backbone of rough set rule induction. It has been shown previously that this property can be maintained for various fuzzy rough set models, including those based on ordered weighted average (OWA) operators. In this paper, we will focus on some instances of the general class of fuzzy quantifier-based fuzzy rough sets (FQFRS). In these models, the lower and upper approximations are evaluated using binary and unary fuzzy quantifiers, respectively. One of the main targets of this study is to examine the granular representation of different models of FQFRS. The main findings reveal that Choquet-based fuzzy rough sets can be represented granularly under the same conditions as OWA-based fuzzy rough sets, whereas Sugeno-based FRS can always be represented granularly. This observation highlights the potential of these models for resolving data inconsistencies and managing noise.
    \end{abstract}


\end{frontmatter}
\section{Introduction}
Fuzzy rough sets (FRS) \cite{dubois1990rough} represent a fusion of fuzzy sets and rough sets, specifically designed to manage vague and potentially inconsistent information. Fuzzy sets model vague information by acknowledging that membership in certain concepts or the logical truth of particular propositions exists on a spectrum. The other part, rough sets, address potential inconsistencies by offering both a lower and upper approximation of a concept concerning an indiscernibility relation between objects.

The criteria for inclusion in the lower and upper approximations within rough sets can be formulated using quantifiers. Traditionally, an object is part of the lower approximation of a concept if all objects indiscernible from it are also members of the concept. In the context of fuzzy quantifier-based fuzzy rough sets (FQFRS) \cite{theerens2022fedcsis,theerens2023fqfrs}, a departure from the conventional universal quantifier is made by employing a fuzzy quantifier such as ``most'', instead of the universal quantifier. This relaxation introduces a degree of tolerance towards noise into the approximations, enhancing their robustness.

Granular representations, whether in the context of rough sets \cite{yao1999rough} or fuzzy rough sets \cite{degang2011granular}, involve expressing the lower and upper approximations as a combination of elementary (fuzzy) sets, referred to as granules, which are derived from the underlying data. These representations hold particular significance in the domain of rule induction. Rule induction entails the generation of a set of rules that establish relationships between object descriptions and decision classes. The granules,
constituting rough sets and fuzzy rough sets, can be interpreted as ``if..., then...'' rules, which are easily comprehensible and interpretable. These granules can be leveraged to construct a rule-based inference system that serves as a predictive model. For instance, the LEM2 algorithm, as outlined in \cite{grzymala1992lers}, is an example of a rule induction algorithm designed for fuzzy rough sets.

In \cite{palangetic2022granular} the authors proved that OWA-based fuzzy rough approximations are granularly representable sets when using D-convex left-continuous t-norms and their residual implicators for calculating the approximations. In this paper, we extend this result to Choquet-based FRS \cite{theerens2022choquet} and show that Sugeno-based FRS are granularly representable without the D-convex condition on the t-norm. Furthermore, we show that several other FQFRS models are not granularly representable, therefore making them less suitable for rule induction purposes.

The following outline is used for the rest of this paper: Section \ref{sec: Preliminaries} provides a review of the necessary preliminaries on FQFRS and the granular representation of OWA-based FRS. Section \ref{sec: granular Choquet} demonstrates that Choquet-based FRS can be represented granularly under the same conditions as OWA-based FRS. Section \ref{sec: granular Sugeno} proves that Sugeno-based fuzzy rough sets can be represented granularly without any extra conditions on the fuzzy set connectives. In Section \ref{sec: non granular FQFRS} the granularity of other FQFRS models is discussed. Finally, Section \ref{sec: conclusion and future work} concludes this paper and outlines future work.

\section{Preliminaries}
\label{sec: Preliminaries}
\subsection{Implicator-conjunctor based fuzzy rough sets}
We will use the notation \(\pwset(X)\) to represent the powerset of \(X\) and assume \(X\) to be finite throughout this paper. Likewise, we will use the notation \(\fpwset(X)\) to represent the set consisting of all fuzzy sets on \(X\).

A fuzzy relation \(R\in \fpwset(X\times X)\) may satisfy one or more of the following properties:
\begin{itemize}
    \item \emph{Reflexivity}: for every \(x\) in \(X\), \(R(x,x)=1\),
    \item \emph{Symmetry}: for every \(x\) and \(y\) in \(X\), \(R(x,y) = R(y,x)\),
    \item \emph{\(\tnorm\)-transitivity} with respect to a t-norm \(\tnorm\): if \(\tnorm(R(x,y),R(y,z))\leq R(x,z)\) holds for every \(x\), \(y\), and \(z\) in \(X\).
\end{itemize}
A fuzzy relation that is reflexive, symmetric, and \(\tnorm\)-transitive is called a \(\tnorm\)-equivalence relation.

We define the \(R\)-foreset of an element \(y\in X\) and a fuzzy relation \(R \in \fpwset(X\times X)\) as the fuzzy set \(Ry(x) := R(x,y)\).

The extension of a mapping \(\mathcal{O}:[0,1]^2 \to [0,1]\) to fuzzy sets (i.e., \(\fpwset(X)^2 \to \fpwset(X)\)) will be denoted by the same symbol:
\[\mathcal{O}(A,B)(x) := \mathcal{O}\left(A(x), B(x)\right), \; \; \forall x \in X.\]

Pawlak \cite{pawlak1982rough} introduced the \emph{lower} and \emph{upper approximation} of $A\in \pwset(X)$ w.r.t.\ an equivalence relation \(R\in \pwset(X\times X)\) as:
\begin{align*}
    \underline{apr}_{R} A & = \left\{\left.x\in X\right| \left[x\right]_R \subseteq A \right\}=\left\{\left.x\in X\right| (\forall y\in X) \left((x,y)\in R \implies y\in A\right)\right\},        \\
    \overline{apr}_{R} A  & = \left\{\left.x\in X\right| \left[x\right]_R \cap A \neq \emptyset \right\}=\left\{\left.x\in X\right| (\exists y\in X) \left((x,y)\in R \land y\in A\right)\right\}.
\end{align*}

In Radzikowska et al.'s work \cite{radzikowska2002comparative}, an implicator-conjunctor-based extension was introduced for fuzzy relations and fuzzy sets. This extension defines the  \emph{lower} and \emph{upper approximation} of $A\in\fpwset(X)$ w.r.t.\ $R\in \fpwset(X\times X)$ as follows:
\begin{align*}
    (\underline{\text{apr}}^{\imp}_{R} A)(x)  & = \min\limits_{y\in X} \imp(R(x,y),A(y)),        \\
    (\overline{\text{apr}}^{\tnorm}_{R} A)(x) & = \max\limits_{y\in X} \mathcal{C}(R(x,y),A(y)),
\end{align*}
where $\imp$ is an implicator\footnote{An \emph{implicator} is a binary operator $\imp: \left[0,1\right]^2\rightarrow \left[0,1\right]$ that is non-increasing in the first argument, non-decreasing in the second argument and for which $\imp(0,0)=\imp(0,1)=\imp(1,1)=1$ and \(\imp(1,0)=0\) holds.} and $\mathcal{C}$ a conjunctor\footnote{A \emph{conjunctor} is a binary operator \(\mathcal{C}:[0,1]^2\to [0,1]\) which is increasing in both arguments, satisfies \(\mathcal{C}(0,0)=\mathcal{C}(0,1)=0\) and for which \(\mathcal{C}(1,x)=x\) holds for all \(x\in[0,1]\). A commutative and associative conjunctor \(\tnorm\) is called a \emph{t-norm}.}.

\begin{prop}[\cite{klement2013triangular}]
    \label{prop: implicator tnorm to imp imp}
    If \(\tnorm\) is a left-continuous t-norm and \(\imp\) its R-implicator\footnote{The residual implicator (R-implicator) of a t-norm \(\tnorm\) is defined as
        \[I_\tnorm (x,y) = \sup\{\lambda \in [0,1] | \tnorm(x,\lambda)\leq y\},\]
        for all \(x,y \in [0,1]\).}, we have
    \[\imp(\tnorm(x,y), z)= \imp(x,\imp(y,z)),\]
    for all \(x,y,z \in [0,1]\).
\end{prop}
\begin{prop}[\cite{d2015comprehensive}]
    \label{prop:exact and idempotence}
    Suppose \(\tnorm\) is a left-continuous t-norm, \(\imp\) its R-implicator, \(R\) a fuzzy \(\tnorm\)-equivalence relation on \(X\) and \(A\in\fpwset(X)\), then the lower and upper approximation satisfy the following properties:
    \begin{itemize}
        \item (\bf{inclusion}) \((\underline{\text{apr}}^\imp_{R} A)\subseteq A \subseteq (\overline{\text{apr}}^\tnorm_{R} A)\),
        \item (\bf{idempotence}) \((\underline{\text{apr}}^\imp_{R} (\underline{\text{apr}}^\imp_{R} A))=(\underline{\text{apr}}^\imp_{R} A), (\overline{\text{apr}}^\tnorm_{R} (\overline{\text{apr}}^\tnorm_{R} A))=(\overline{\text{apr}}^\tnorm_{R} A)\),
        \item (\bf{exact approximation}) \(\forall A\in\fpwset(X):(\underline{\text{apr}}^\imp_{R} A) = A  \Leftrightarrow A=(\overline{\text{apr}}^\tnorm_{R} A)\).
    \end{itemize}
\end{prop}

The extension of the concept of granular representability for fuzzy rough approximations was first explored by Degang et al. in 2011 \cite{degang2011granular}. Here, the authors introduced the notion of a fuzzy granule as:
\[R_\lambda(x) = \{(y, \tnorm(\lambda, R(x, y))) \;|\; y\in X\},\]
where, \(\lambda\) ranges in the interval \([0,1]\), \(x\) belongs to the set \(X\), and \(\tnorm\) represents a t-norm.

\begin{defn}{\cite{degang2011granular}}
    We call \(A\in\fpwset(X)\) granularly representable if
    \[A = \bigcup\left\{R_\lambda(x) | \lambda \in [0,1], x\in X, R_\lambda(x) \subseteq  A\right\}\]
\end{defn}
\begin{prop}{\cite{degang2011granular}}
    \label{prop: granularfuzzy}
    Let \(\tnorm\) be a left-continuous t-norm, \(\imp\) its residual implicator and \(R\) a \(\tnorm\)-equivalence relation on \(X\).
    A fuzzy set \(A\in\fpwset(X)\) is granularly representable w.r.t.\ the relation \(R\) if and only if \(\underline{\text{apr}}^{\imp}_R (A) = A = \overline{\text{apr}}^{\tnorm}_R(A)\).
\end{prop}
The intuition behind this definition is that if \(A\) can be constructed using simple sets, i.e., fuzzy granules, then \(A\) will be free from any inconsistencies.
\begin{prop}{\cite{degang2011granular}}
    \label{prop: consistentfuzzy}
    A fuzzy set \(A\in\fpwset(X)\) is granularly representable if and only if it satisfies the consistency property, i.e.,
    \[\tnorm(R(x,y),A(y)) \leq A(x),\; \forall x,y \in X.\]
\end{prop}

\begin{defn}
    An element \(y\) is called consistent with respect to a fuzzy relation \(R\in \mathcal{P}(X \times X)\) and a fuzzy set \(A\in\fpwset(X)\) if and only if
    \[\tnorm(R(x,y),A(y))\leq A(x), \qquad\forall x\in X.\]
\end{defn}
\begin{prop}
    \label{inconsistencyInOneElement}
    Let \(\tnorm\) be a t-norm and \(\imp\) its residual implicator. An element \(y\) is consistent with respect to a reflexive fuzzy relation \(R\in \mathcal{P}(X \times X)\) and a fuzzy set \(A\in\fpwset(X)\) if and only if \(\underline{apr}^\imp_{R}(A)(y) = A(y)\).
\end{prop}
\begin{proof}
    Since \(\imp\) is the residual implicator of \(\tnorm\), we have
    \begin{align*}
        y \text{ consistent} & \Leftrightarrow (\forall x\in X)(\tnorm(R(x,y),A(y))\leq A(x)) \\
                             & \Leftrightarrow (\forall x\in X)(A(y)\leq \imp(R(x,y),A(x)))   \\
                             & \Leftrightarrow A(y) \leq \inf_{x\in X}\imp(R(x,y),A(x))       \\
                             & \Leftrightarrow A(y) = \inf_{x\in X}\imp(R(x,y),A(x))          \\
                             & \Leftrightarrow A(y) = \underline{apr}_{R}(A)(y),
    \end{align*}
    where in the second-to-last step, we used the reflexivity of the relation and the property that, for residual implicators, \(\imp(1,x) = x\) holds for all \(x \in X\).
\end{proof}

\subsection{Choquet and Sugeno integral}
Choquet and Sugeno integrals extend the concept of classical integration to a context where the measures are not necessarily additive. This allows for more flexible and realistic modeling of uncertainty and imprecision in data, making them useful in a wide range of applications, most notably decision-making \cite{grabisch2010decade}.
\begin{defn}
    A set function \(\mu:\mathcal{P}(X)\to[0,1]\) is a \emph{monotone measure} if:
    \begin{itemize}
        \item $\mu(\emptyset)=0$ and \(\mu(X)=1\),
        \item \((\forall A,B\in\mathcal{P}(X))(A\subseteq B \implies \mu(A)\leq \mu(B))\).
    \end{itemize}
    A monotone measure is \emph{symmetric} when \(\abs{A}=\abs{B}\) implies \(\mu(A)=\mu(B)\).
\end{defn}

\begin{defn}[\cite{wang2010generalized}]
    \label{defn: ChoquetIntegral}
    The \emph{Choquet integral} of \(f:X\to\mathbb{R}\) with respect to a monotone measure $\mu$ on \(X\) is defined as:
    \begin{equation*}
        \int f \diff \mu=\sum_{i=1}^n f(x^\ast_i)\cdot\left[\mu(A^\ast_i)-\mu(A^\ast_{i+1})\right],
    \end{equation*}
    where \((x^\ast_1,x^\ast_2,\dots,x^\ast_n)\) is a permutation of \(X=(x_1,x_2,\dots,x_n)\) such that
    \begin{equation*}
        f(x^\ast_1)\leq f(x^\ast_2) \leq\cdots\leq f(x^\ast_n),
    \end{equation*}
    \(A^\ast_i:=\{x^\ast_i,\dots,x^\ast_n\}\) and \(\mu(A^\ast_{n+1}):=0\).
\end{defn}

\begin{defn}[\cite{wang2010generalized}]
    The \emph{Sugeno integral} of \(f:X\to\mathbb{R}\) with respect to a monotone measure $\mu$ on \(X\) is defined as:
    \begin{equation*}
        \fint f \diff \mu=\max_{i=1}^n \min\left(\mu\left(\{x^\ast_i,\dots, x^\ast_n\}\right), f\left(x^*_i\right)\right),
    \end{equation*}
    where \((x^\ast_1,x^\ast_2,\dots,x^\ast_n)\) is a permutation of \(X=(x_1,x_2,\dots,x_n)\) such that
    \begin{equation*}
        f(x^\ast_1)\leq f(x^\ast_2) \leq\cdots\leq f(x^\ast_n).
    \end{equation*}
\end{defn}
We recall that Ordered Weighted Average \cite{yagerOWA} operators are equivalent to Choquet integrals w.r.t.\ symmetric monotone measures \cite{beliakov2007aggregation}.

\subsection{Fuzzy quantifier-based fuzzy rough sets}
One class of robust (i.e.,\ noise tolerant) FRS is based on fuzzy quantifiers \cite{theerens2023fqfrs}. For a thorough exposition of the theory of fuzzy quantifiers, we refer the reader to \cite{glockner2008fuzzy}.
\begin{defn}[\cite{glockner2008fuzzy}]
    An \(n\)-ary semi-fuzzy quantifier on \(X\neq \emptyset\) is a mapping \(Q:(\pwset(X))^n\to [0,1]\). An \(n\)-ary fuzzy quantifier on \(X\neq \emptyset\) is a mapping \(\widetilde{Q}:\left(\fpwset(X)\right)^n\to [0,1]\).
\end{defn}
\begin{defn}[\cite{glockner2008fuzzy}]
    Let \(\widetilde{Q}:\left(\fpwset(X)\right)^2\to [0,1]\) be a binary quantification model over the universe \(X\), then the corresponding unary quantification model \(Unary(\widetilde{Q}):\fpwset(X)\to [0,1]\) is defined as \(Unary(\widetilde{Q})(A):= \widetilde{Q}(X,A)\), \(A\in\fpwset(X)\).
\end{defn}

\begin{defn}[\cite{theerens2023fqfrs}]
    Given a reflexive fuzzy relation \(R\in\fpwset(X\times X)\), fuzzy quantifiers \(\widetilde{Q}_l:(\fpwset(X))^2\to [0,1]\) and \(\widetilde{Q}_u:\fpwset(X)\to [0,1]\), and $A\in\fpwset(X)$, then the \emph{lower} and \emph{upper approximation} of $A$ w.r.t.\ $R$ are given by:
    \begin{align*}
        (\underline{apr}_{R,\widetilde{Q}_l}A)(y) & =\widetilde{Q}_l\left(Ry,A \right),       \\
        (\overline{apr}_{R,\widetilde{Q}_u}A)(y)  & =\widetilde{Q}_u(Ry\cap_{\mathcal{T}} A),
    \end{align*}
\end{defn}
Let \(\widetilde{Q}_l\) and \(\widetilde{Q}_u\) denote the (linguistic) quantifiers ``almost all'' and ``some'' respectively. Then the membership degree of an element \(y\) to the lower approximation of \(A\) corresponds to the truth value of the statement ``Almost all elements similar to \(y\) are in \(A\)''. Similarly, the membership degree of \(y\) in the upper approximation is determined by the truth value of the statement ``Some elements are similar to \(y\) and are in \(A\)''. Note that there exists a distinction in the quantification models applied to the lower and upper approximations. The upper approximation involves unary quantification, as the proposition ``\(Q\) elements are in \(A\) and \(B\)'' fundamentally represents a unary proposition with the fuzzy set \(\tnorm(A,B)\) as its argument. This contrasts with the lower approximation, where the proposition ``\(Q\) \(A\)'s are \(B\)'s'' serves as the underlying proposition, employing a necessarily binary quantification model.

To specify a quantifier like ``some'' on general universes we will make use of RIM quantifiers.
\begin{defn}[\cite{yager1996quantifier}]
    A fuzzy set \(\Lambda\in\fpwset([0,1])\) is called a \emph{regular increasing monotone (RIM) quantifier} if
    \(\Lambda\) is a non-decreasing function such that \(\Lambda(0)=0\) and \(\Lambda(1)=1\).
\end{defn}
The interpretation of the RIM quantifier \(\Lambda\) is that if \(p\) is the percentage of elements for which a certain proposition \(P\) holds, then \(\Lambda(p)\) determines the truth value of the quantified proposition \(\Lambda P\).

Suppose \(\mu_l\) and \(\mu_u\) are monotone measures on a finite universe \(X\) and \(\Lambda_l\) and \(\Lambda_u\) are two RIM-quantifiers, then we can define the following fuzzy quantifier fuzzy rough sets (FQFRS).
\begin{itemize}
    \item When we define \(\widetilde{Q}_l = C^\imp_{\mu_l}\) and \(\widetilde{Q}_u = Unary(C^\imp_{\mu_u})\), where
          \[C^\imp_\mu(A,B) = \int \imp(A,B)\diff \mu,\]
          and restricting \(\mu_l\) and \(\mu_u\) to symmetric measures, we get OWA-based fuzzy rough sets (OWAFRS) \cite{cornelis2010ordered}. When we allow general monotone measures, we get Choquet-based fuzzy rough sets (CFRS) \cite{theerens2022choquet}. By permitting non-symmetry in \(\mu\), we increase our flexibility to incorporate additional information from the dataset (cf.\ \cite{theerens2022choquet}).
    \item When we define \(\widetilde{Q}_l = S^\imp_{\mu_l}\) and \(\widetilde{Q}_u = Unary(S^\imp_{\mu_u})\), where
          \[S^\imp_\mu(A,B) = \fint \imp(A,B)\diff \mu,\]
          we get Sugeno-based fuzzy rough sets (SFRS).
    \item Let \(\widetilde{Q}_l = \widetilde{YWIC}_{\Lambda_l}\) and \(\widetilde{Q}_u = Unary(\widetilde{YWIC}_{\Lambda_u})\), where
          \begin{align}
              \label{ywic}
              \widetilde{YWIC}_\Lambda (A,B) & = \int \imp(A,B)\diff\mu^\ast_{A},\;\;\mu^\ast_A(E)=\Lambda\left(\frac
              {\sum_{i=1}^{\abs*{E}}A(y^\ast_i)}{\abs{A}}\right),
          \end{align}
          where \(y^\ast_i\) is defined such that \(C(y^\ast_i)\) is the \(i\)th smallest value of \(C(x)\) for all \(x\in X\) and \(i \in \{1,\dots,\abs{X}=n\}\). Then the FQFRS corresponding to these quantifiers is YWIC-FQFRS \cite{theerens2023fqfrs}, replacing the Choquet-integral with a Sugeno-integral we get YWIS-FQFRS.
    \item  When we define \(\widetilde{Q}_l = \widetilde{WC}_{\Lambda_l}\) and \(\widetilde{Q}_u = Unary(\widetilde{WC}_{\Lambda_u})\), where
          \begin{align}
              \label{wowac}
              \widetilde{WC}_\Lambda (A,B) & = \int \imp(A,B)\diff\mu_{A},\;\;\mu_A(E)=\Lambda\left(\frac
              {\abs{A\cap E}}{\abs{A}}\right),
          \end{align}
          we get WOWAC-FQFRS \cite{theerens2023fqfrs}, replacing the Choquet-integral with a Sugeno-integral we get WOWAS-FQFRS.
\end{itemize}
Note that \(Unary(\widetilde{YWIC}_{\Lambda})\) and \(Unary(\widetilde{WC}_{\Lambda})\) are the same and are equal to
\[\widetilde{Y}_\Lambda(A) = \int A \diff \mu_\Lambda, \;\; \mu_\Lambda(E)=\Lambda\left(\frac
    {\abs{E}}{\abs{X}}\right).\]
Because \(\mu_\Lambda\) represents a general symmetric measure, we have that \(\widetilde{Y}_\Lambda\) is an OWA operator (Yager's OWA model for fuzzy quantification, cf.\ \cite{yager1996quantifier}), hence the upper approximations of YWIC- and WOWAC-FQFRS are equivalent to those of OWAFRS. For a comparison between some of these different lower approximations, we refer the reader to \cite{theerens2023fqfrs}.

In \cite{palangetic2022granular}, the authors showed that for a specific type of fuzzy connectives and for a \(\tnorm\)-equivalence relation, OWA-based fuzzy rough approximations do not
possess inconsistencies, i.e., they are granularly representable fuzzy sets.
\begin{defn}[\cite{matouvsek2001directional}]
    We say that a binary operator \(H: [0,1]^2 \to [0,1] \) is directionally convex or D-convex (directionally concave or D-concave) if it is a convex (concave) function in both of its arguments, i.e., for all \(x_1,x_2, y\in[0,1]\) and \(w_1,w_2\in[0,1]\) such that \(w_1+w_2=1\), it holds that:
    \[H(w_1x_1+w_2x_2,y)\leq(\geq)\:w_1H(x_1,y)+w_2H(x_2,y) ,\] \[H (y , w_1 x_1 + w_2 x_2 ) \leq (\geq) \:w_1 H (y , x_1 ) + w_2 H (y , x_2).\]
\end{defn}

\begin{prop}[\cite{palangetic2022granular}]
    \label{Implicator_concave}
    Let \(\tnorm\) be a D-convex left-continuous t-norm and \(\imp\) its R-implicator. Then \(\imp\) is concave in its second argument.
\end{prop}

\begin{prop}[\cite{palangetic2022granular}]
    Let \(\tnorm\) be a D-convex left-continuous t-norm, \(\imp\) its residual implicator, \(\mathbf{w}_u\) and \(\mathbf{w}_l\) two weight vectors and \(R\) a \(\tnorm\)-equivalence relation. Then for every \(A \in \fpwset(X)\) we have
    \[\underline{apr}^\imp_{R}(\underline{apr}_{R, \mathbf{w}_l}(A)) = \underline{apr}_{R, \mathbf{w}_l}(A) \text{ and }\;\overline{apr}^\tnorm_{R}(\overline{apr}_{R, \mathbf{w}_u}(A)) = \overline{apr}_{R, \mathbf{w}_u}(A),\]
    where \(\underline{apr}_{R, \mathbf{w}_l}\) and \(\overline{apr}_{R, \mathbf{w}_u}\) denote the OWA-based lower and upper approximation, respectively.
\end{prop}
\begin{cor}[\cite{palangetic2022granular}]
    Let \(\tnorm\) be a D-convex left-continuous t-norm, \(\imp\) its residual implicator, \(\mathbf{w}_u\) and \(\mathbf{w}_l\) two weight vectors and \(R\) a \(\tnorm\)-equivalence relation. Then for every \(A \in \fpwset(X)\) we have
    \begin{align*}
        \underline{apr}_{R, \mathbf{w}_l}(A) & = \bigcup\left\{R_\lambda(x) | R_\lambda(x) \subseteq  \underline{apr}_{R, \mathbf{w}_l}(A)\right\} \\
        \overline{apr}_{R, \mathbf{w}_u}(A)  & = \bigcup\left\{R_\lambda(x) | R_\lambda(x) \subseteq  \overline{apr}_{R, \mathbf{w}_u}(A)\right\},
    \end{align*}
    where \(\underline{apr}_{R, \mathbf{w}_l}\) and \(\overline{apr}_{R, \mathbf{w}_u}\) denote the OWA-based lower and upper approximation, respectively.
\end{cor}

\section{Granularity of Choquet-based fuzzy rough sets}\
\label{sec: granular Choquet}
In this section, we generalize the granular representation of OWA-based fuzzy rough sets (cf.\ \cite{palangetic2022granular}) to Choquet-based fuzzy rough sets. In particular, we show that under the same requirements on the fuzzy connectives, the Choquet-based fuzzy rough approximations are still free from any inconsistencies, i.e.,\ they are granularly representable sets. We first recall Jensen's inequality \cite{jensen1906fonctions}, and then prove a specific variant for Choquet integrals.
\begin{prop}[Jensen's inequality]
    Let \(f:\mathbb{R}\to\mathbb{R}\) be a convex (concave) function, \(x_1,x_2,\dots,x_n\in\mathbb{R}\) and \(w_1,w_2,\dots,w_n\in[0,1]\) weights (\(\sum_{i=1}^n w_i = 1\)). Then we have
    \[f\left(\sum_{i=1}^n w_i \cdot x_i\right) \leq\; (\geq) \sum_{i=1}^n w_i \cdot f(x_i).\]
\end{prop}
\begin{lem}[Jensen's inequality for Choquet integrals]
    Let \(\Phi: \mathbb{R}^+ \to \mathbb{R}^+\) be a non-decreasing function, \(f: X \to \mathbb{R}^+\) and \(\mu\) a monotone measure on \(X\). If \(\Phi\) is convex, we have
    \[\Phi\left(\int f \diff \mu\right) \leq \int \Phi(f) \diff \mu.\]
    If \(\Phi\) is concave, we have
    \[\Phi\left(\int f \diff \mu\right) \geq \int \Phi(f) \diff \mu.\]
\end{lem}
\begin{proof}
    Let \(\Phi\) be convex, the proof for a concave \(\Phi\) proceeds analogously.
    Using Jensen's inequality we have
    \begin{align}
        \Phi \left(\int f \diff \mu\right) & =\Phi\left(\sum_{i=1}^{n}f(x^\ast_i)\cdot \left[\mu(A^\ast_i) - \mu(A^\ast_{i+1})\right]\right) \nonumber                              \\
                                           & \leq \sum_{i=1}^{n}\Phi(f(x^\ast_i))\cdot \left[\mu(A^\ast_i) - \mu(A^\ast_{i+1})\right]= \int \Phi(f) \diff \mu, \label{lasteqjensen}
    \end{align}
    where we can apply Jensen's inequality because of
    \[\sum_{i=1}^n \left[\mu(A^\ast_i) - \mu(A^\ast_{i+1})
            \right] =\mu(X)- \mu(\emptyset) = 1,\] and the last equality in Eq.\ \eqref{lasteqjensen} holds because of the non-decreasingness of \(\Phi\) (hence order-preserving).
\end{proof}

\begin{lem}
    \label{min_max_integral}
    Let \(\mu\) be a monotone measure on \(X\) and \(f:X^2\to \mathbb
    R\). Then we have the following inequalities
    \begin{align*}
        \int \min_{x\in X} f(x,y) \diff \mu & \leq \min_{x\in X} \int f(x,y)\diff \mu(y), \\
        \int \max_{x\in X} f(x,y) \diff \mu & \geq \max_{x\in X} \int f(x,y)\diff \mu(y),
    \end{align*}
    where \(\int\) denotes either a Choquet integral or Sugeno integral.
\end{lem}
\begin{proof}
    Follows directly from the increasingness of the Choquet and Sugeno-integrals and the fact that \[\min_{z\in X} f(z,y) \leq f(x,y) \text{ and } \max_{z\in X} f(z,y) \geq f(x,y),\]
    for every \(x\in X\).
\end{proof}
\begin{thm}
    \label{granularity_Choquet}
    Let \(\tnorm\) be a D-convex left-continuous t-norm, \(\imp\) its residual implicator, \(\mu_u\) and \(\mu_l\) two monotone measures and \(R\) a \(\tnorm\)-equivalence relation. Then for every \(A \in \fpwset(X)\) we have
    \[\underline{apr}_{R}(\underline{apr}^{\mu_l}_R(A)) = \underline{apr}^{\mu_l}_R(A) \text{ and }\;\overline{apr}_{R}(\overline{apr}^{\mu_u}_R(A)) = \overline{apr}^{\mu_u}_R(A),\]
    where \(\underline{apr}^{\mu_l}_R\) and \(\overline{apr}^{\mu_u}_R\) denote the Choquet lower and upper approximation, respectively.
\end{thm}
\begin{proof}
    Observe that because of the \(\tnorm\)-transitivity and reflexivity we have
    \[R(x,y) = \max_{z\in X}\tnorm(R(x,z),(z,y)).\]
    We start with  the lower approximation. Note that
    \[\underline{apr}_{R}(\underline{apr}^{\mu_l}_R(A)) \subseteq \underline{apr}^{\mu_l}_R(A)\]
    follows directly from the inclusion property of the lower approximation.
    Using this and noting that \(\imp\) is concave in its second argument (Proposition \ref{Implicator_concave}), thus allowing the use of Jensen's inequality for Choquet integrals, we obtain the other inclusion:
    \begin{align*}
        \underline{apr}^{\mu_l}_R(A)(y) & = \int \imp (Ry,A) \diff\mu_l                                                                                                             \\
                                        & = \int \imp\left(\max_{z\in X}\tnorm\left(R(x,z),R(z,y)\right), A(x)\right)\diff \mu_l(x)                                                 \\
                                        & = \int \min_{z\in X}\imp\left(\tnorm\left(R(x,z),R(z,y)\right), A(x)\right)\diff \mu_l(x)                                                 \\
                                        & \leq \min_{z\in X}\int \imp\left(\tnorm\left(R(x,z),R(z,y)\right), A(x)\right)\diff \mu_l(x)                                              \\
                                        & = \min_{z\in X}\int \imp\left(R(z,y), \imp\left( R(x,z),A(x)\right)\right)\diff \mu_l(x)                                                  \\
                                        & \leq \min_{z\in X} \imp\left(R(z,y), \int\imp\left( R(x,z),A(x)\right)\diff \mu_l(x)\right)                                               \\
                                        & = \min_{z\in X} \imp\left(R(z,y), \underline{apr}^{\mu_l}_R(A)(z)\right)=\underline{apr}_{R}\left(\underline{apr}^{\mu_l}_R(A)\right)(y),
    \end{align*}
    where we made use of the monotonicity of t-norms and implicators, Proposition \ref{prop: implicator tnorm to imp imp}, Lemma \ref{min_max_integral} and Jensen's inequality for Choquet integrals with \(\Phi(x)= \imp(R(z,y),x)\).
    For the upper approximations the proof proceeds analogously:
    \begin{align*}
        \overline{apr}^{\mu_u}_R(A)(y) & = \int \tnorm(Ry,A) \diff\mu_u                                                                                                           \\
                                       & = \int \tnorm\left(\max_{z\in X}\tnorm\left(R(x,z),R(z,y)\right), A(x)\right)\diff \mu_u(x)                                              \\
                                       & = \int \max_{z\in X}\tnorm\left(\tnorm\left(R(x,z),R(z,y)\right), A(x)\right)\diff \mu_u(x)                                              \\
                                       & \geq \max_{z\in X}\int \tnorm\left(\tnorm\left(R(x,z),R(z,y)\right), A(x)\right)\diff \mu_u(x)                                           \\
                                       & = \max_{z\in X}\int \tnorm\left(R(z,y), \tnorm\left( R(x,z),A(x)\right)\right)\diff \mu_u(x)                                             \\
                                       & \geq \max_{z\in X} \tnorm\left(R(z,y), \int\tnorm\left( R(x,z),A(x)\right)\diff \mu_u(x)\right)                                          \\
                                       & = \max_{z\in X} \tnorm\left(R(z,y), \overline{apr}^{\mu_u}_R(A)(z)\right)=\overline{apr}_{R}\left(\overline{apr}^{\mu_u}_R(A)\right)(y),
    \end{align*}
    where commutativity and associativity of the t-norm is used as well as Lemma \ref{min_max_integral} and the Jensen inequality for Choquet integrals with \(\Phi(x)= \tnorm(R(z,y),x)\).
    The other inclusion follows directly from the inclusion property of upper approximations.
\end{proof}
\begin{cor}
    Let \(\tnorm\) be a D-convex left-continuous t-norm, \(\imp\) its residual implicator, \(\mu_u\) and \(\mu_l\) two monotone measures and \(R\) a \(\tnorm\)-equivalence relation. Then for every \(A \in \fpwset(X)\) we have
    \begin{align*}
        \underline{apr}^{\mu_l}_{R}(A) & = \bigcup\left\{R_\lambda(x) | R_\lambda(x) \subseteq  \underline{apr}^{\mu_l}_{R}(A)\right\} \\
        \overline{apr}^{\mu_u}_{R}(A)  & = \bigcup\left\{R_\lambda(x) | R_\lambda(x) \subseteq  \overline{apr}^{\mu_u}_{R}(A)\right\},
    \end{align*}
    where \(\underline{apr}^{\mu_l}_R\) and \(\overline{apr}^{\mu_u}_R\) denote the Choquet lower and upper approximation, respectively.
\end{cor}
\begin{proof}
    Directly follows from Proposition \ref{prop: granularfuzzy} and the exact approximation property of fuzzy rough sets (Proposition \ref{prop:exact and idempotence}).
\end{proof}

\section{Granularity of Sugeno-based fuzzy rough sets}
\label{sec: granular Sugeno}
In this section, we prove that Sugeno-based fuzzy rough sets are granularly representable under the same conditions as classical fuzzy rough sets. As a result, they are free from any inconsistencies. Sugeno-based lower and upper approximations can thus be seen as a way to simultaneously remove inconsistencies and noise.
\begin{lem}[Jensen's inequality for Sugeno integrals]
    \label{lem: Jensen Sugeno}
    Let \(\mu\) be a monotone measure on a finite universe \(X\), \(f: X \to \mathbb{R}^+\)  and \(\Phi: \mathbb{R}^+ \to \mathbb{R}^+\) be a non-decreasing function. If \(\Phi(x)\geq x\), for every \(x\in [0,\mu(X)]\), we have
    \[\Phi\left( \fint f \diff \mu\right) \geq \fint \Phi(f) \diff \mu.\]
    If \(\Phi(x)\leq x\), for every \(x\in [0,\mu(X)]\), we have
    \[\Phi\left(\fint f \diff \mu\right) \leq\fint \Phi(f) \diff \mu.\]
\end{lem}
\begin{proof}
    We will only prove the first inequality, the second inequality is proved analogously. Let \((x^\ast_1,x^\ast_2,\dots,x^\ast_n)\) be a permutation of \(X=(x_1,x_2,\dots,x_n)\) such that
    \begin{equation*}
        f(x^\ast_1)\leq f(x^\ast_2) \leq\cdots\leq f(x^\ast_n),
    \end{equation*}
    and define \(A^\ast_i := \{x^\ast_i,\cdots,x^\ast_n\}\).
    Note that because of the non-decreasingness of \(\Phi\) we have
    \begin{equation*}
        \Phi(f(x^\ast_1)) \leq \Phi(f(x^\ast_2)) \leq\cdots\leq \Phi(f(x^\ast_n)),
    \end{equation*}
    hence
    \begin{align*}
        \Phi \left(\fint f \diff \mu\right) & = \Phi\left(\max_{i=1}^n \min\left(\mu\left(A^\ast_i\right), f(x^\ast_i)\right)\right)                  \\
                                            & = \max_{i=1}^n \left(\Phi\left(\min\left(\mu\left(A^\ast_i\right), f(x^\ast_i)\right)\right)\right)     \\
                                            & = \max_{i=1}^n \min\left(\Phi\left(\mu\left(A^\ast_i\right)\right), \Phi\left(f(x^\ast_i)\right)\right) \\
                                            & \geq \max_{i=1}^n \min\left(\mu\left(A^\ast_i\right), \Phi\left(f(x^\ast_i)\right)\right)               \\
                                            & = \fint\Phi(f)\diff \mu,
    \end{align*}
    where the first inequality follows from the fact that \(\Phi(x)\geq x\) for every \(x\in [0,\mu(X)]\).
\end{proof}
\begin{thm}
    Let \(\tnorm\) be a left-continuous t-norm, \(\imp\) its residual implicator, \(\mu_u\) and \(\mu_l\) two monotone measures and \(R\) a \(\tnorm\)-equivalence relation. Then for every \(A \in \fpwset(X)\) we have
    \[\underline{apr}_{R}(\underline{apr}^{\mu_l}_R(A)) = \underline{apr}^{\mu_l}_R(A) \text{ and }\;\overline{apr}_{R}(\overline{apr}^{\mu_u}_R(A)) = \overline{apr}^{\mu_u}_R(A),\]
    where \(\underline{apr}^{\mu_l}_R\) and \(\overline{apr}^{\mu_u}_R\) denote the Sugeno lower and upper approximation, respectively.
\end{thm}
\begin{proof}
    For the lower approximation, note that \(\Phi(x)= \imp (R(z,y),x)\) satisfies the requirements of Lemma \ref{lem: Jensen Sugeno}, i.e.,\ \(\Phi(x)\geq x\) and increasing. Indeed,
    \[x \leq \max \{\lambda : \tnorm(y, \lambda) \leq x\}=\imp(y,x),\]
    because of \(\tnorm(y,x)\leq \min(y,x)\leq x\),
    while the increasingness follows from the increasingness of \(\imp\) in the second argument. The rest of the proof is analogous to that of Theorem \ref{granularity_Choquet}. For the upper approximation, note that \(\Phi(x)= \tnorm (R(z,y),x)\) satisfies the requirements of Lemma \ref{lem: Jensen Sugeno}, i.e.,\ \(\Phi(x)\leq x\) and increasing. Indeed,
    \[x \geq \min(y,x) \geq \tnorm(y,x),\]
    while the increasingness follows from the increasingness of \(\tnorm\). The rest of the proof is analogous to that of Theorem \ref{granularity_Choquet}.
\end{proof}

\begin{cor}
    Let \(\tnorm\) be a left-continuous t-norm, \(\imp\) its residual implicator, \(\mu_u\) and \(\mu_l\) two monotone measures and \(R\) a \(\tnorm\)-equivalence relation. Then for every \(A \in \fpwset(X)\) we have
    \begin{align*}
        \underline{apr}^{\mu_l}_{R}(A) & = \bigcup\left\{R_\lambda(x) | R_\lambda(x) \subseteq  \underline{apr}^{\mu_l}_{R}(A)\right\} \\
        \overline{apr}^{\mu_u}_{R}(A)  & = \bigcup\left\{R_\lambda(x) | R_\lambda(x) \subseteq  \overline{apr}^{\mu_u}_{R}(A)\right\},
    \end{align*}
    where \(\underline{apr}^{\mu_l}_R\) and \(\overline{apr}^{\mu_u}_R\) denote the Sugeno lower and upper approximation, respectively.
\end{cor}
\begin{proof}
    Directly follows from Proposition \ref{prop: granularfuzzy} and the exact approximation property of fuzzy rough sets (Proposition \ref{prop:exact and idempotence}).
\end{proof}

\section{Granularity of other fuzzy quantifier based fuzzy rough sets}
\label{sec: non granular FQFRS}
In this section, we will show that YWI-FQRFRS and WOWA-FQFRS are not granularly representable. In addition, we will show that on realistic datasets these inconsistencies do not occur frequently.
\subsection{Counterexamples of the granularity of YWI-FQFRS and WOWA-FQFRS}
The following examples show that YWI-FQFRS and WOWA-FQFRS (both the Choquet and Sugeno versions) are not granularly representable, even when using a convex t-norm and its residual implicator. We will make use of the following well-known propositions.
\begin{prop}
    The \L ukasiewicz t-norm \(\tnorm_L(x,y)=\max(0,x+y-1)\) is convex.
\end{prop}
Also note that the R-implicator of the \L ukasiewicz t-norm \(\tnorm_L(x,y)=\max(0,x+y-1)\) is equal to the \L ukasiewicz implicator \(\imp_L(x,y)=\min(1,1-x+y)\).
\begin{example}[Counterexample for the granularity of YWIC-FQFRS]\mbox{}\\
    Let \(X = \{x_1,x_2,x_3,x_4,x_5\}\), \(\Lambda\) the identity RIM-quantifier (\(\Lambda(x)= x\)) and \(A = \{x_1,x_2,x_3\}\), which we will also denote as \[A = [1.0, 1.0, 1.0, 0.0, 0.0].\] Furthermore, suppose we have one attribute \(a\) on \(X\) that is given by \([1.0, 0.5, 1.0, 0.0, 0.0]\). Using this attribute and the fuzzy \(\tnorm_L\)-equivalence relation \(R\) on \(X\) defined by
    \[R(x,y)=1-\abs{a(y)-a(x)},\]
    we get the following membership degrees:
    \begin{equation*}
        R = \begin{bmatrix}
            1.0 & 0.5 & 1.0 & 0.0 & 0.0 \\
            0.5 & 1.0 & 0.5 & 0.5 & 0.5 \\
            1.0 & 0.5 & 1.0 & 0.0 & 0.0 \\
            0.0 & 0.5 & 0.0 & 1.0 & 1.0 \\
            0.0 & 0.5 & 0.0 & 1.0 & 1.0
        \end{bmatrix}.
    \end{equation*}
    Making use of \(\Lambda(x)=x\), we have that Eq.\ \eqref{ywic} reduces to
    \[\widetilde{YWIC}_\Lambda (C,B)  = \sum_{i=1}^n (\imp_L(C,B))(x^{\ast}_i)\cdot \frac{C(y^\ast_i)}{\abs{C}}.\]
    Let us now calculate \(\underline{apr}^\text{YWIC}_{R}(A)\):
    \begin{align*}
        \underline{apr}^\text{YWIC}_{R}(A)(x) = \widetilde{YWIC}_\Lambda (Rx,A) = \sum_{i=1}^5 (\imp_L(Rx,A))(x^{\ast}_i)\cdot \frac{Rx(y^\ast_i)}{\abs{Rx}},
    \end{align*}
    where \(x^\ast_i\) and \(y^\ast_i\) are defined such that \(\imp(Rx,A)(x^{\ast}_i)\) is the \(i\)th largest value of \(\imp(Rx,A)(y)\) and \(Rx(y^\ast_i)\) is the \(i\)th smallest value of \(Rx(y)\) for all \(y\in X\) and \(i \in \{1,\dots,5\}\).
    Calculating \(\imp_L(Rx,A)\) (\(\imp_L(x,y) = \min(1,1-x+y)\)) gives us
    \begin{align*}
        [\imp_L(R(x_i,x_j), A(x_j))] = \begin{bmatrix}
                                           1.0 & 1.0 & 1.0 & 1.0 & 1.0 \\
                                           1.0 & 1.0 & 1.0 & 0.5 & 0.5 \\
                                           1.0 & 1.0 & 1.0 & 1.0 & 1.0 \\
                                           1.0 & 1.0 & 1.0 & 0.0 & 0.0 \\
                                           1.0 & 1.0 & 1.0 & 0.0 & 0.0
                                       \end{bmatrix}.
    \end{align*}
    Notice that this is already sorted, so \(x^\ast_i = x_i\) for \(i \in \{1,\dots,5\}\). We now sort \(R\):
    \begin{align*}
        [R(x_i,(y^\ast_i)_j)] = \begin{bmatrix}
                                    0.0 & 0.0 & 0.5 & 1.0 & 1.0 \\
                                    0.5 & 0.5 & 0.5 & 0.5 & 1.0 \\
                                    0.0 & 0.0 & 0.5 & 1.0 & 1.0 \\
                                    0.0 & 0.0 & 0.5 & 1.0 & 1.0 \\
                                    0.0 & 0.0 & 0.5 & 1.0 & 1.0
                                \end{bmatrix}.
    \end{align*}
    Adding everything together, we get (\(\abs{Rx} = [2.5, 3.0, 2.5,2.5,2.5]\))
    \[\underline{apr}^\text{YWIC}_{R}(A)= [1.0, 0.75, 1.0, 0.2, 0.2].\]
    Let us now calculate \(\underline{apr}_{R}(\underline{apr}^\text{YWIC}_{R}(A))\):
    \begin{align*}
        \imp_L\left(R(x_i,x_j),\underline{apr}^\text{YWIC}_{R}(A)(x_j)\right) = \begin{bmatrix}
                                                                                    1.0 & 1.0  & 1.0 & 1.0 & 1.0 \\
                                                                                    1.0 & 0.75 & 1.0 & 0.7 & 0.7 \\
                                                                                    1.0 & 1.0  & 1.0 & 1.0 & 1.0 \\
                                                                                    1.0 & 1.0  & 1.0 & 0.2 & 0.2 \\
                                                                                    1.0 & 1.0  & 1.0 & 0.2 & 0.2
                                                                                \end{bmatrix},
    \end{align*}
    thus
    \[\underline{apr}_{R}(\underline{apr}^\text{YWIC}_{R}(A)) = [1.0, 0.7, 1.0, 0.2, 0.2] \neq \underline{apr}^\text{YWIC}_{R}(A),\]
    and
    \[\underline{apr}^\text{YWIC}_{R}(A) - \underline{apr}_{R}(\underline{apr}^\text{YWIC}_{R}(A))  = [0.0, 0.05, 0.0, 0.0, 0.0].\]
\end{example}
\begin{example}[Counterexample for the granularity of YWIS-FQFRS]\mbox{}\\
    Let \(X = \{x_1,x_2,x_3,x_4,x_5\}\), \(\Lambda\) the identity RIM-quantifier (\(\Lambda(x)= x\)) and \(A = \{x_1,x_2,x_3\}\), which we will also denote as \[A = [1.0, 1.0, 1.0, 0.0, 0.0].\] Furthermore, suppose we have one attribute \(a\) on \(X\) that is given by \([0, 0, 0.2, 0.91, 1]\). Using this attribute and the fuzzy \(\tnorm_L\)-equivalence relation \(R\) on \(X\) defined by
    \[R(x,y)=1-\abs{a(y)-a(x)},\]
    we get the following membership degrees:
    \begin{equation*}
        R = \begin{bmatrix}
            1.0  & 1.0  & 0.8  & 0.09 & 0.0  \\
            1.0  & 1.0  & 0.8  & 0.09 & 0.0  \\
            0.8  & 0.8  & 1.0  & 0.29 & 0.2  \\
            0.09 & 0.09 & 0.29 & 1.0  & 0.91 \\
            0.0  & 0.0  & 0.2  & 0.91 & 1.0
        \end{bmatrix}.
    \end{equation*}

    Through a straightforward calculation we get:
    \[\underline{apr}^\text{YWIS}_{R}(A)=  [0.91, 0.91, 0.71, 0.19747, 0.0948],\]
    \[\underline{apr}_{R}(\underline{apr}^\text{YWIS}_{R}(A)) = [0.91, 0.91, 0.71, 0.18479, 0.0948],\]
    and
    \[\underline{apr}^\text{YWIS}_{R}(A) - \underline{apr}_{R}(\underline{apr}^\text{YWIS}_{R}(A))  = [0.0, 0.0, 0.0, 0.01267, 0.0].\]
\end{example}

\begin{example}[Counterexample for the granularity of WOWAC-FQFRS]\mbox{}\\
    Let \(X = \{x_1,x_2,x_3\}\), \(\Lambda\) the identity RIM-quantifier (\(\Lambda(x)= x\)) and \(A = \{x_2\}\), which we will also denote as \[A = [0.0, 1.0, 0.0].\] Furthermore, suppose we have one attribute \(a\) on \(X\) that is given by \([0.0, 0.5, 0.0]\). Using this attribute and the fuzzy \(\tnorm_L\)-equivalence relation \(R\) on \(X\) defined by
    \[R(x,y)=1-\abs{a(y)-a(x)},\]
    we get the following membership degrees:
    \begin{equation*}
        R = \begin{bmatrix}
            1.0 & 0.5 & 1.0 \\
            0.5 & 1.0 & 0.5 \\
            1.0 & 0.5 & 1.0
        \end{bmatrix}.
    \end{equation*}
    Making use of \(\Lambda(x)=x\), we have that Eq.\ \eqref{wowac} reduces to
    \[\widetilde{W}^{\imp}_\Lambda (C,B)  = \sum_{i=1}^n (\imp_L(C,B))(x^{\ast}_i)\cdot \frac{C(x^\ast_i)}{\abs{C}}=\sum_{i=1}^n (\imp_L(C,B))(x_i)\cdot \frac{C(x_i)}{\abs{C}}.\]
    Let us now calculate \(\underline{apr}^\text{WOWAC}_{R}(A)\):
    \begin{align*}
        \underline{apr}^\text{WOWAC}_{R}(A)(x) = \widetilde{W}^\imp_\Lambda (Rx,A) = \sum_{i=1}^5 (\imp_L(Rx,A))(x_i)\cdot \frac{Rx(x_i)}{\abs{Rx}}.
    \end{align*}
    Calculating \(\imp_L(Rx,A)\) (\(\imp_L(x,y) = \min(1,1-x+y)\)) gives us
    \begin{align*}
        [\imp_L(R(x_i,x_j), A(x_j))] = \begin{bmatrix}
                                           0.0 & 1.0 & 0.0 \\
                                           0.5 & 1.0 & 0.5 \\
                                           0.0 & 1.0 & 0.0
                                       \end{bmatrix}.
    \end{align*}
    Adding everything together, we get (\(\abs{Rx} = [2.5, 2, 2.5]\))
    \[\underline{apr}^\text{WOWAC}_{R}(A)= [0.2, 0.75,0.2].\]
    Let us now calculate \(\underline{apr}_{R}(\underline{apr}^\text{WOWAC}_{R}(A))\):
    \begin{align*}
        \imp_L\left(R(x_i,x_j),\underline{apr}^\text{WOWAC}_{R}(A)(x_j)\right) = \begin{bmatrix}
                                                                                     0.2 & 1.0  & 0.2 \\
                                                                                     0.7 & 0.75 & 0.7 \\
                                                                                     0.2 & 1.0  & 0.2
                                                                                 \end{bmatrix} ,
    \end{align*}
    thus
    \[\underline{apr}_{R}(\underline{apr}^\text{WOWAC}_{R}(A)) = [0.2, 0.7, 0.2]\neq \underline{apr}^\text{WOWAC}_{R}(A),\]
    and
    \[\underline{apr}^\text{WOWAC}_{R}(A) - \underline{apr}_{R}(\underline{apr}^\text{WOWAC}_{R}(A))  = [0.0, 0.05, 0.0].\]
\end{example}
\begin{example}[Counterexample for the granularity of WOWAS-FQFRS]\mbox{}\\
    Let \(X = \{x_1,x_2,x_3,x_4,x_5\}\), \(\Lambda\) the identity RIM-quantifier (\(\Lambda(x)= x\)) and \(A = \{x_1,x_2,x_3\}\), which we will also denote as \[A = [1.0, 1.0, 0.5, 0.5, 0.0].\] Furthermore, suppose we have one attribute \(a\) on \(X\) that is given by \([0.8, 0.0, 0.0, 0.0, 1.0]\). Using this attribute and the fuzzy \(\tnorm_L\)-equivalence relation \(R\) on \(X\) defined by
    \[R(x,y)=1-\abs{a(y)-a(x)},\]
    we get the following membership degrees:
    \begin{equation*}
        R = \begin{bmatrix}
            1.0 & 0.2 & 0.2 & 0.2 & 0.8 \\
            0.2 & 1.0 & 1.0 & 1.0 & 0.0 \\
            0.2 & 1.0 & 1.0 & 1.0 & 0.0 \\
            0.2 & 1.0 & 1.0 & 1.0 & 0.0 \\
            0.8 & 0.0 & 0.0 & 0.0 & 1.0
        \end{bmatrix}.
    \end{equation*}

    Through a straightforward calculation we get:
    \[\underline{apr}^\text{WOWAS}_{R}(A)=  [0.666\dots, 0.5, 0.5, 0.5, 0.44\dots],\]
    \[\underline{apr}_{R}(\underline{apr}^\text{WOWAS}_{R}(A)) = [0.644\dots, 0.5, 0.5, 0.5, 0.44\dots],\]
    and
    \[\underline{apr}^\text{WOWAS}_{R}(A) - \underline{apr}_{R}(\underline{apr}^\text{WOWAS}_{R}(A))  = [0.022\dots, 0.0, 0.0, 0.0, 0.0].\]
\end{example}
As we can see in the above counterexamples, the inconsistencies always occur only in one element (cf.\ Proposition \ref{inconsistencyInOneElement}), and the difference between the FQFRS approximations and their inconsistency-free lower approximations never exceeds 0.05. In the next section, we will evaluate if this observation also extends to typical real-life datasets.
\subsection{Granularity on realistic datasets}
Because the WOWA and YWI FQFRS models lack granular representability, inconsistencies may arise in their lower approximations, rendering them unsuitable for rule induction. Yet, the question remains: to what extent does this issue manifest in real-world datasets?  To answer this question we will examine the occurrence of these inconsistencies in classification datasets.
\subsubsection{Setup}
To measure the extent of inconsistencies still remaining in the lower approximation of a dataset, we will compute the following error as well as the percentage of inconsistent elements (cf.\ Proposition \ref{inconsistencyInOneElement}):\
\[\frac{\sum_{C\in d}|\underline{apr}^\text{FQFRS}_{R}(C) - \underline{apr}_{R}(\underline{apr}^\text{FQFRS}_{R}(C))|}{\text{\#Classes}*\text{\#Instances}},\]
and
\[\frac{\sum_{C\in d}|\{x\in X: \underline{apr}^\text{FQFRS}_{R}(C)(x) - \underline{apr}_{R}(\underline{apr}^\text{FQFRS}_{R}(C)(x)>0)\}|}{\text{\#Classes}*\text{\#Instances}},\]
where \(d\) is the set of classes, and \(\text{\#Classes}\) and \(\text{\#Instances}\) represent the number of classes and instances, respectively. This calculation is based on Proposition \ref{prop: granularfuzzy} and the exact approximation property of fuzzy rough sets (as shown in Proposition \ref{prop:exact and idempotence}). When there are no inconsistencies, this value will be zero.
We will conduct our experiment on 20 classification datasets (Table \ref{dataset_description}) from the UCI repository \cite{Dua:2019}. The different FQFRS models we will evaluate are:
\[\text{FQFRS} \in \{\text{WOWAC}, \text{WOWAS}, \text{YWIC}, \text{YWIS}\}.\]
The \L ukasiewicz t-norm and its residuated implicator are used in all models.
The lower approximations will be evaluated using the RIM quantifiers \(\Lambda(x) = \Lambda_{(\alpha,1)}(x)\) with
\begin{align*}
    \label{ZADEH_S_function}
    \Lambda_{(\alpha, \beta)}(p) & = \left\{
    \begin{array}{ll}
        0                                       & \;p\leq \alpha                               \\
        \frac{2(p-\alpha)^2}{(\beta-\alpha)^2}  & \; \alpha \leq p \leq \frac{\alpha+\beta}{2} \\
        1-\frac{2(p-\beta)^2}{(\beta-\alpha)^2} & \;  \frac{\alpha+\beta}{2}\leq p\leq \beta   \\
        1                                       & \; \beta\leq p
    \end{array}
    \right.,
\end{align*}
Zadeh's S-function \cite{cornelis2007vaguely} and
\[\alpha \in  \{0.6,0.7,\dots,0.9,0.91,0.92,\dots,0.99,1\},\]
where we have chosen finer steps at the end to observe the convergence of \(\Lambda_{(\alpha,1)}\) to the universal quantifier as \(\alpha\) approaches \(1\). We will use the following \(\tnorm_L\)-equivalence relation:
\begin{align*}
     & R(x,y)=\sum_{a\in\mathcal{A}}\left[\max\left(0,1-\frac{\abs{a(y)-a(x)}}{\sigma_a}\right)\right],
\end{align*}
where \(\mathcal{A}\) is the set of conditional attributes and \(\sigma_a\) denotes the standard deviation of a conditional attribute \(a\in \mathcal{A}\).
\begin{table}[H]
    \small
    \begin{center}
        \begin{tabular}{l|c|c|c||l|c|c|c}

            Name         & \# Ft. & \# Inst. & \# Cl. & Name           & \# Ft. & \# Inst. & \# Cl. \\
            \hline
            accent       & 12     & 329      & 6      & ionosphere     & 34     & 351      & 2      \\
            \hline
            appendicitis & 7      & 106      & 2      & leaf           & 14     & 340      & 30     \\
            \hline
            banknote     & 4      & 1372     & 2      & pop-failures   & 18     & 540      & 2      \\
            \hline
            biodeg       & 41     & 1055     & 2      & segment        & 19     & 2310     & 7      \\
            \hline
            breasttissue & 9      & 106      & 6      & somerville     & 6      & 143      & 2      \\
            \hline
            coimbra      & 9      & 116      & 2      & sonar          & 60     & 208      & 2      \\
            \hline
            debrecen     & 19     & 1151     & 2      & spectf         & 44     & 267      & 2      \\
            \hline
            faults       & 27     & 1941     & 7      & sportsarticles & 59     & 1000     & 2      \\
            \hline
            haberman     & 3      & 306      & 2      & transfusion    & 4      & 748      & 2      \\
            \hline
            ilpd         & 10     & 579      & 2      & wdbc           & 30     & 569      & 2      \\
        \end{tabular}
    \end{center}
    \caption{Description of the 20 used UCI datasets (\# Cl. = number of classes, \# Ft.\ = number of features, \# Inst.\ = number of instances).}
    \label{dataset_description}
\end{table}
\subsubsection{Results and discussion}
\begin{figure}[H]
    \begin{center}
        \includegraphics[scale=0.335]{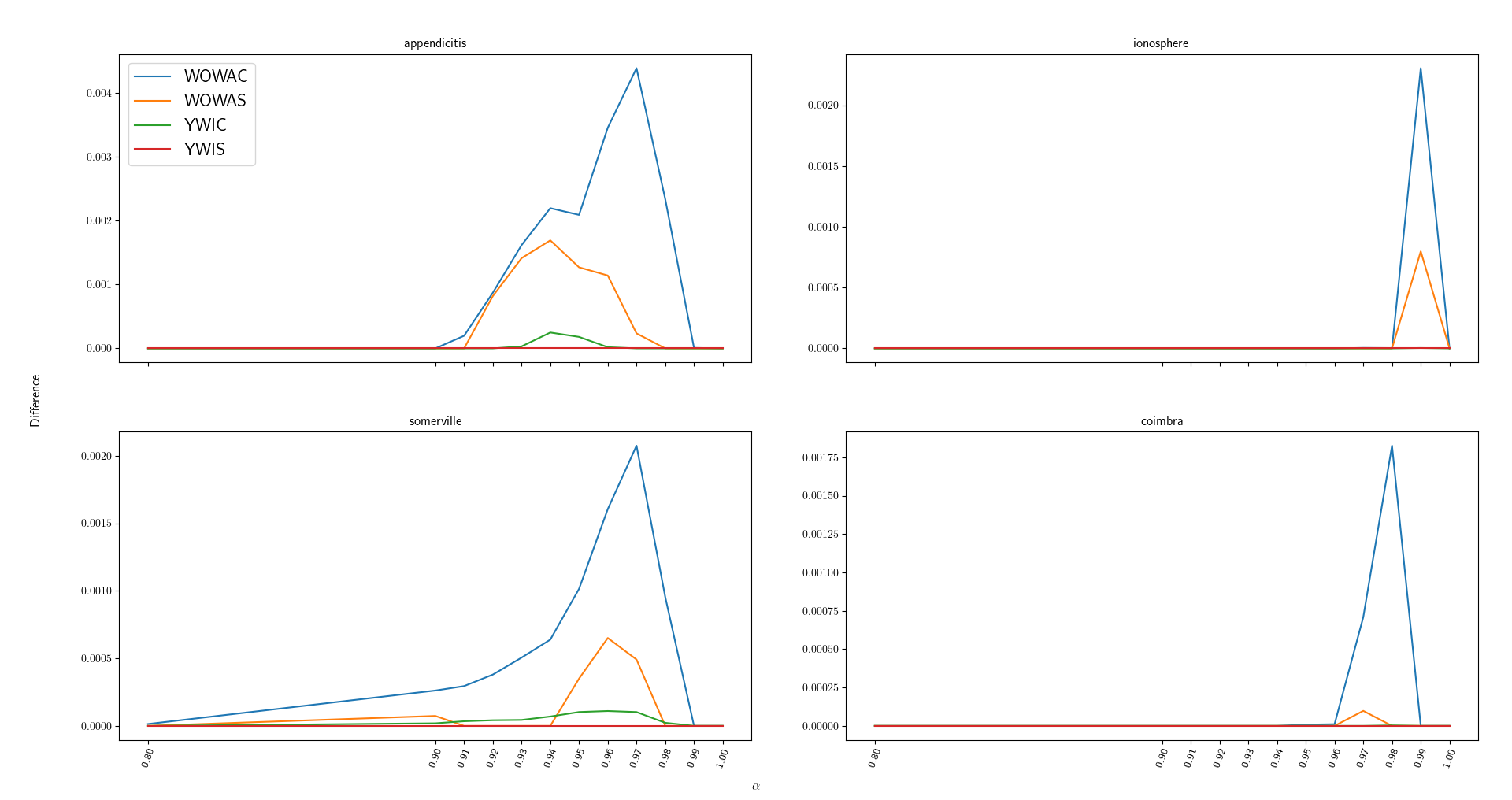}
    \end{center}
    \caption{Plot of the error with respect to the \(\alpha\)-parameter for the 4 datasets with the highest maximal error.}
    \label{errorPlot}
\end{figure}

\begin{figure}[H]
    \begin{center}
        \includegraphics[scale=0.335]{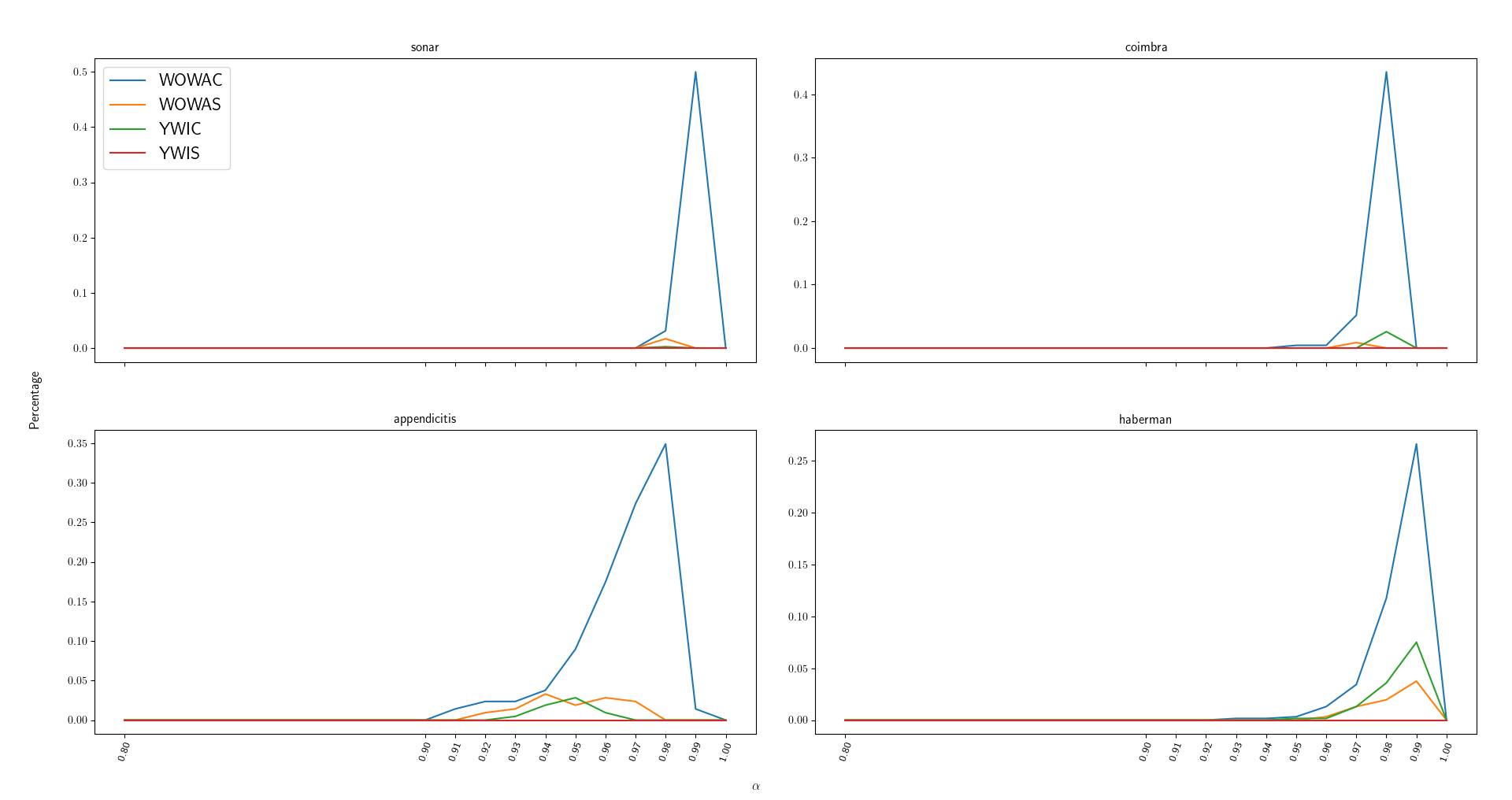}
    \end{center}
    \caption{Plot of the percentage of inconsistencies with respect to the \(\alpha\)-parameter for the 4 datasets with the highest maximal percentage of inconsistencies.}
    \label{percentagePlot}
\end{figure}

Table \ref{maximalErrorTable} shows the maximal error and maximal percentage of inconsistencies over all models and \(\alpha\)-values for each dataset. This table reveals that the errors do not exceed 0.004, and for certain datasets, they are negligible, since they can almost certainly be attributed to floating-point errors. The WOWAC model attains nearly all of these maximum errors. Table \ref{maximalErrorTable YWIC} illustrates the maximum error and the highest percentage of inconsistencies across all \(\alpha\)-values for the YWIC model. Note the substantial contrast between the two models, with the YWIC model exhibiting fewer inconsistencies compared to the WOWAC model.
\begin{table}[H]
    \begin{center}
        \begin{tabular}{c | c | c || c | c | c}
            Dataset        & Error    & Percentage & Dataset      & Error    & Percentage \\
            \hline
            ilpd           & 0        & 0          & transfusion  & 3.47e-04 & 6.95e-02   \\
            sportsarticles & 0        & 0          & accent       & 3.53e-04 & 4.36e-02   \\
            faults         & 1.63e-20 & 1.47e-04   & banknote     & 5.22e-04 & 6.16e-02   \\
            debrecen       & 1.93e-19 & 4.34e-04   & sonar        & 1.36e-03 & 5.00e-01   \\
            biodeg         & 3.31e-09 & 2.37e-03   & haberman     & 1.37e-03 & 2.68e-01   \\
            segment        & 4.09e-08 & 1.67e-03   & spectf       & 1.55e-03 & 2.40e-01   \\
            leaf           & 3.44e-05 & 7.65e-03   & coimbra      & 1.83e-03 & 4.35e-01   \\
            wdbc           & 3.92e-05 & 2.64e-03   & somerville   & 2.07e-03 & 2.10e-01   \\
            breasttissue   & 2.94e-04 & 9.28e-02   & ionosphere   & 2.30e-03 & 1.32e-01   \\
            pop-failures   & 3.23e-04 & 2.41e-02   & appendicitis & 4.39e-03 & 3.49e-01   \\
        \end{tabular}
    \end{center}
    \caption{Maximal error and maximal percentage of inconsistencies over all models and \(\alpha\)-values.}
    \label{maximalErrorTable}
\end{table}
\begin{table}[H]
    \begin{center}
        \begin{tabular}{c | c | c || c | c | c}
            Dataset        & Error    & Percentage & Dataset      & Error    & Percentage \\
            \hline
            biodeg         & 0        & 0          & accent       & 8.43e-07 & 4.56e-03   \\
            debrecen       & 0        & 0          & coimbra      & 2.99e-06 & 2.59e-02   \\
            faults         & 0        & 0          & ionosphere   & 3.05e-06 & 1.28e-02   \\
            ilpd           & 0        & 0          & wdbc         & 7.83e-06 & 8.79e-04   \\
            pop-failures   & 0        & 0          & breasttissue & 1.08e-05 & 7.86e-03   \\
            sportsarticles & 0        & 0          & banknote     & 1.52e-05 & 1.31e-02   \\
            segment        & 9.07e-09 & 1.24e-04   & transfusion  & 2.13e-05 & 1.60e-02   \\
            sonar          & 1.15e-07 & 2.40e-03   & somerville   & 1.10e-04 & 4.55e-02   \\
            leaf           & 1.74e-07 & 7.84e-04   & haberman     & 1.10e-04 & 7.68e-02   \\
            spectf         & 5.08e-07 & 7.49e-03   & appendicitis & 2.49e-04 & 2.83e-02   \\
        \end{tabular}
    \end{center}
    \caption{Maximal error and maximal percentage of inconsistencies over all \(\alpha\)-values for the YWIC model.}
    \label{maximalErrorTable YWIC}
\end{table}
To further examine the differences between the four FQFRS models, we depict the error with respect to the \(\alpha\)-parameter for the top four datasets with the maximum errors in Figure \ref{errorPlot} and the percentage of inconsistencies for the top four datasets with the maximum percentage of inconsistencies in Figure \ref{percentagePlot}. In all cases, the WOWA models consistently display the most significant errors and a higher level of inconsistencies, with the Choquet variant (WOWAC) occupying the highest position. For the YWI models, the Choquet variant also displays the highest error, whereas the Sugeno variant (YWIS) exhibits no inconsistent elements or errors largely originating from floating-point errors.

\section{Conclusion and future work}
\label{sec: conclusion and future work}
In summary, this paper explored the granular representation of fuzzy quantifier-based fuzzy rough sets (FQFRS). We established that Choquet-based FQFRS can be granularly represented under the same conditions as OWA-based fuzzy rough sets, while Sugeno-based FQFRS can be granularly represented under the same conditions as classical fuzzy rough sets. This discovery highlights the potential of these models for resolving data inconsistencies and managing noise.

Additionally, we examined models that incorporate extra weighting on the first argument, such as WOWA and YWI. Our findings indicated that these models do not yield granularly representable lower approximations. However, it is worth noting that in practical situations, these approaches still demonstrate effectiveness in mitigating inconsistencies, as demonstrated in our experiments.

Looking ahead, it remains an open question whether there exist FQFRS models that introduce an extra weighting on the first argument while still achieving granularly representable lower approximations. Furthermore, there is room for exploration into the possibility of achieving granular representation under more relaxed conditions, such as extra conditions on the t-norm or through the development of weaker versions of granular representability.
\bibliography{/Users/adnantheerens/Desktop/phd/Articles/bibfiles/bibfile}
\end{document}